\documentclass[10pt,twocolumn,letterpaper]{article}

\usepackage{cvpr}              % To produce the CAMERA-READY version
\usepackage{graphicx}
\usepackage{amsmath}
\usepackage{amssymb}
\usepackage{booktabs}
\usepackage{multirow}
\usepackage{makecell} % For multirowcell

\usepackage{amsfonts,bm}
\newcommand{\R}{\mathbb{R}}
\usepackage[pagebackref,breaklinks,colorlinks]{hyperref}

\graphicspath{ {figs/} }
\usepackage[capitalize]{cleveref}
\crefname{section}{Sec.}{Secs.}
\Crefname{section}{Section}{Sections}
\Crefname{table}{Table}{Tables}
\crefname{table}{Tab.}{Tabs.}

\def\cvprPaperID{6820} % *** Enter the CVPR Paper ID here
\def\confName{CVPR}
\def\confYear{2023}

\newcommand{\figref}[1]{Fig.~\ref{#1}}

\newcommand{\tabref}[1]{Tab.~\ref{#1}}
\newcommand{\apref}[1]{\ref*{#1}}
\newcommand{\sdf}{\mathsf{S}}
\newcommand{\mlp}{\text{MLP}}
\newcommand{\pe}{\text{PE}}
\newcommand{\Loss}{\mathcal{L}}
\newcommand{\Lreg}{\Loss_\text{reg}}
\newcommand{\Lcolor}{\Loss_\text{color}}

\newcommand{\modelname}{PET-NeuS}
\newcommand{\tpbackbone}{TP-NeuS}
\begin{document}

\title{PET-NeuS: Positional Encoding Tri-Planes for Neural Surfaces}
\date{}
\author{Yiqun Wang \\ Chongqing University and KAUST
\and
Ivan Skorokhodov \\ KAUST
\and
Peter Wonka \\ KAUST
}
\maketitle

%%%%%%%%% ABSTRACT
\begin{abstract}
A signed distance function (SDF) parametrized by an MLP is a common ingredient of neural surface reconstruction. We build on the successful recent method NeuS to extend it by three new components. The first component is to borrow the tri-plane representation from EG3D and represent signed distance fields as a mixture of tri-planes and MLPs instead of representing it with MLPs only. Using tri-planes leads to a more expressive data structure but will also introduce noise in the reconstructed surface. The second component is to use a new type of positional encoding with learnable weights to combat noise in the reconstruction process. We divide the features in the tri-plane into multiple frequency scales and modulate them with sin and cos functions of different frequencies. The third component is to use learnable convolution operations on the tri-plane features using self-attention convolution to produce features with different frequency bands. The experiments show that PET-NeuS achieves high-fidelity surface reconstruction on standard datasets. Following previous work and using the Chamfer metric as the most important way to measure surface reconstruction quality, we are able to improve upon the NeuS baseline by 57\% on Nerf-synthetic (0.84 compared to 1.97) and by 15.5\% on DTU (0.71 compared to 0.84). The qualitative evaluation reveals how our method can better control the interference of high-frequency noise. Code available at \url{https://github.com/yiqun-wang/PET-NeuS}.
\end{abstract}

%%%%%%%%% BODY TEXT
\section{Introduction}
\label{sec:intro}

Implicit neural functions, or neural fields, have received a lot of attention in recent research. The seminal paper NeRF~\cite{Nerf} combines neural fields with volume rendering, enabling high-quality novel view synthesis. Inspired by NeRF, NeuS~\cite{NeuS} and VolSDF~\cite{Volsdf} introduce a signed distance function (SDF) into the volume rendering equation and regularize the SDF, so that smooth surface models can be reconstructed. However, these methods use pure MLP networks to encode SDFs. Although these two methods can reconstruct smooth surfaces, they both leave room for improvement when it comes to reconstructing surface details.

\begin{figure*}[!t]
\begin{center}
\includegraphics[width=1.0\textwidth]{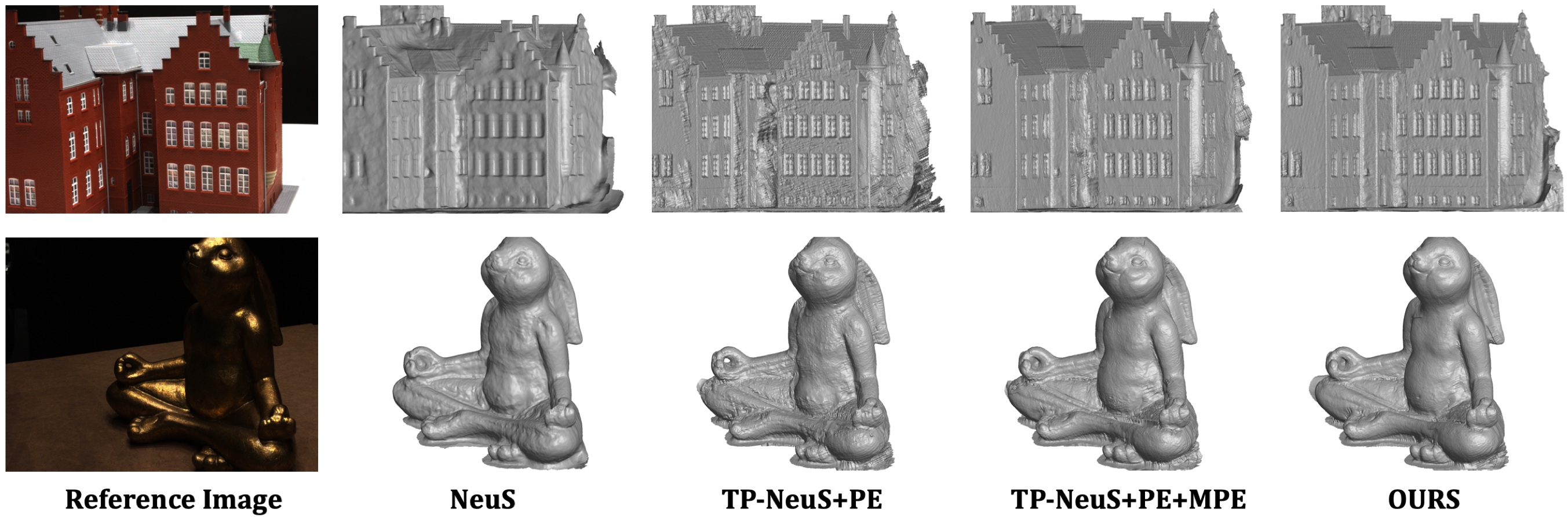}
\end{center}
\caption{The challenge of using the tri-plane representation directly. First column: reference image. Second to the fifth column: NeuS, Learning SDF using tri-planes, OURS without self-attention convolution, and OURS.}
\label{fig:MA}
\end{figure*}

One research direction (\cite{Plenoxels,Kilonerf,Instant-ngp,Tensorf,EG3D}) explores data structures such as tri-planes or voxel grids that are suitable to improve the NeRF framework, in terms of speed or reconstruction quality. However, data structures that are successful for novel view synthesis may not bring immediate success when employed for surface reconstruction as shown in the third column of Fig.~\ref{fig:MA}. While a greater expressiveness to encode local details is useful to better fit the input data, there is also less inductive bias towards a smooth surface. Therefore, noise during image acquisition, high-frequency shading, or high-frequency texture variations are more likely to result in a noisy reconstructed surface.

In our work, we explore how to increase expressiveness to encode local features while at the same time reducing the impact of noise interference. We choose to build on the tri-plane data structure since it consumes less memory and can be easier scaled to higher resolutions.

In our work, we build on EG3D and NeuS to propose a novel framework, called PET-NeuS. First, we propose a method to integrate the tri-plane data structure into a surface reconstruction framework in order to be able to model an SDF with more local details. 
Second, since the features between tri-plane pixels do not share learnable parameters, we use positional encoding to modulate the tri-plane features, thereby enhancing the smoothness of the learnable features. Third, the positional encoding involves functions of different frequencies. In order to better match different frequencies, we propose to use multi-scale self-attention convolution kernels with different window sizes to perform convolution in the spatial domain to generate features of different frequency bands. This further increases the fidelity of the surface  reconstruction while suppressing noise.

We experiment on two datasets to verify the effectiveness of our method, the DTU dataset and the NeRF-Synthetic dataset. Since the DTU dataset contains non-Lambertian surfaces, the ability of the network to resist noise interference can be verified. The NeRF-Synthetic dataset has many sharp features, which can verify that our framework can effectively utilize its improved local expressiveness to better reconstruct local details. We show superior performance compared to state-of-the-art methods on both datasets.

In summary, our contributions are as follows:

\begin{itemize}
    \item We propose to train neural implicit surfaces with a tri-plane architecture to enable the reconstructed surfaces to better preserve fine-grained local features.
    \item We derive a novel positional encoding strategy to be used in conjunction with tri-plane features in order to reduce noise interference.
    \item We utilize self-attention convolution to produce tri-plane features with different frequency bands to match the positional encoding of different frequencies, further improving the fidelity of surface reconstruction.
\end{itemize}

%------------------------------------------------------------------------
\section{Related Work}
\label{sec:related}

\textbf{Neural fields}.
Neural fields are a popular representation of 3D scenes. Two fundamental representations are to encode occupancy ~\cite{Occupancy_net,IM-Net} or signed distance functions~\cite{Deepsdf} using MLPs.
In order to improve the representational expressiveness of the models, \cite{Conv_occupancy,Implicit_Shape} use 3D convolutions on voxels to learn local shape features and construct the occupancy function and signed distance function of the shapes, respectively.
Due to locality, implicit neural function representations can model fine-grained scenes.
Subsequently, some works~\cite{Ini_Sdf,UDF} focus on solving the problem of learning implicit functions on shapes with boundary, while others~\cite{Acorn,NeuralLOD,NeuralSpline} further exploit voxel representations to improve the quality of modeling.
Then the seminal work, NeRF~\cite{Nerf}, incorporates the implicit neural function into the volume rendering formula, thus achieving high-fidelity rendering results.
Due to the implicit neural function representing the scene, the method produces excellent results for novel view synthesis. Some follow-up works~\cite{Siren,Mip,SAPE,Modulated,Refnerf} use multi-scale techniques or encoding strategies to learn fine-grained details. Recently, many works~\cite{Plenoxels,Kilonerf,Instant-ngp,Tensorf,EG3D} use voxel grids or a factored representation (e.g. tri-planes) to further improve training speed or rendering quality.

\begin{figure*}[!t]
\begin{center}
\includegraphics[width=1.0\textwidth]{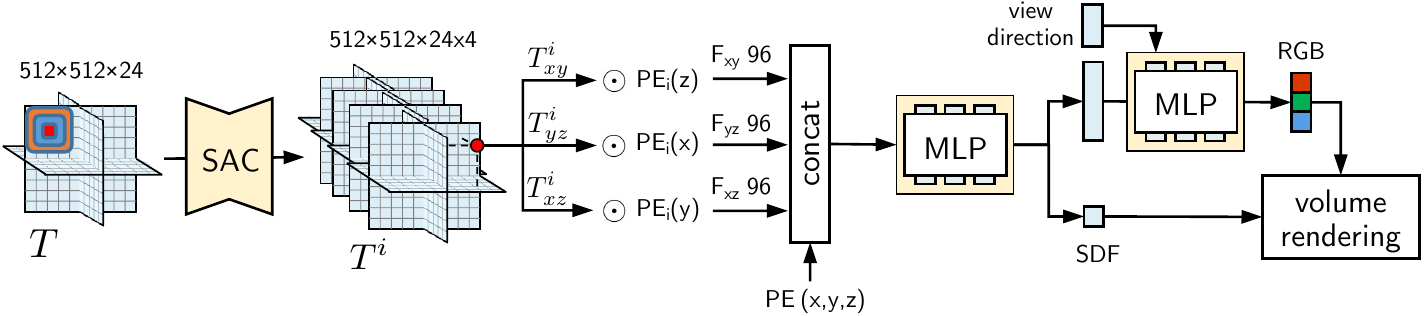}
\end{center}
\vspace{-0.5cm}
\caption{Our PET-NeuS framework consists of a tri-plane architecture, two types of positional encoding, self-attention convolution (SAC), and MLP mapping blocks.}
\label{fig:framework}
\end{figure*}

\textbf{Neural surface reconstruction}.
Surface reconstruction from multiple views is a popular topic in 3D vision. Traditional algorithms for multi-view surface reconstruction usually use discrete voxel-based representations~\cite{Poxels,Photorealistic,Theory,Probabilistic,Kinectfusion,Voxelhashing} or reconstruct point clouds~\cite{Patchmatch,Accurate,Pixelwise,Structure,Gipuma}. Discrete voxel representations suffer from resolution and memory overhead, while point-based methods require additional consideration of missing point clouds and additional surface reconstruction steps. 
Recently, some methods based on neural implicit surfaces have emerged to reconstruct shapes using continuous neural implicit function from multi-view images. Surface rendering and volume rendering are two key techniques.
DVR~\cite{DVR} and IDR~\cite{IDR} adopt surface rendering to model the occupancy functions or signed distance functions for 3D shapes, respectively.  The methods based on surface rendering need to compute the precise location of the surface to render images and gradient descent is applied only on the surface. NeRF-based methods like UNISURF~\cite{Unisurf}, VolSDF~\cite{Volsdf}, and NeuS~\cite{NeuS} incorporate occupancy functions or the signed distance functions into the volume rendering equation. Since the implicit function can be regularized by the Eikonal loss, the reconstructed surface can maintain smoothness. The NeuralPatch method by~\cite{Neuralwarp} is a post-processing step to VolSDF. It binds the colors in the volume to nearby patches with a homography transformation. Since the computation of patch warping relies on accurate surface normals, we consider the algorithm as a post-process that can be applied to any method. 
HF-NeuS~\cite{HF-NeuS} introduces an additional MLP for modeling a displacement field to learn high-frequency details and further improve surface fidelity.  We choose VolSDF, NeuS, and HF-NeuS as our state-of-the-art competitors.

%------------------------------------------------------------------------
\section{Method}
\label{sec:method}

Given a set of images and their camera positions, our goal is to reconstruct the scene geometry represented by a signed distance function (SDF).
In this section, we first provide the necessary details on the tri-plane representation~\cite{Conv_occupancy, EG3D}.
Then, we describe how they can be integrated into NeuS without losing its geometric MLP initialization and, after that, how to combine the grid-based tri-plane representation with conventional sinusoidal positional encoding.
Finally, we introduce multi-frequency tri-plane features.
Our overall framework is illustrated in Fig.~\ref{fig:framework}.

\subsection{Tri-Plane-based NeuS}\label{sec:method:tpbackbone}

A signed distance function (SDF) $\sdf$ is a powerful surface representation suitable for many downstream applications and easily convertible to other representation types~\cite{CG_textbook, Deepsdf}.
It takes a 3D coordinate $(x, y, z) \in \R^3$ as an input, and outputs the (signed) distance $\sdf({x,y,z}) = d_s \in \R$ to the nearest point on the scene surface.
Neural surface (NeuS)~\cite{NeuS} is the most common way to recover such an SDF-based representation of a scene geometry from its (posed) multi-view images.
It uses a multi-layer perceptron (MLP) to model the SDF and optimizes it for scene reconstruction using volumetric rendering~\cite{Nerf}.
The conventional NeuS relies solely on neural networks to encode the scene, which is an expensive representation with limited expressivity~\cite{Siren}.
In this work, we explore tri-planes~\cite{Conv_occupancy, EG3D} as an additional learnable data structure for modeling SDFs (TP-NeuS).

A tri-plane $T$ is a grid-based 3D data structure, which is composed of three learnable feature planes $T = (T_{xy}, T_{yz}, T_{xz})$ of resolution $R \times R$ and feature dimensionality $n_f$ (i.e., $T_{*} \in \R^{R \times R \times n_f}$).
These planes are orthogonal to each other and form a 3D cube of size $L^3$ centered at the origin $(0, 0, 0)$.
To extract a feature $T(x,y,z) = \bm{w} \in \R^{3 n_f}$ at a 3D coordinate $(x, y, z) \in [-\frac{L}{2},\frac{L}{2}]^3$, we project it onto each of the three planes, bilinearly interpolate its nearby feature vectors and concatenate them: $\bm{w} = (\bm{w}_{xy}, \bm{w}_{yz}, \bm{w}_{xz})$.

After computing the feature vector $\bm{w}$, we use it to estimate the signed distance $d_s$.
We do this via a shallow 3-layer MLP: $(d_s, \bm{u}) = \mlp_d(\bm{w}_{xy}, \bm{w}_{yz}, \bm{w}_{xz})$.
In addition to the distance value $d_s$, it produces a feature vector $\bm{u} \in \R^{256}$, which is passed to the color branch.
It is also represented as a 3-layer MLP and predicts view-dependent RGB color value $\mlp_c(\bm{u}, \bm{v}_d) = \text{RGB} \in [0, 1]^3$ from the feature vector $\bm{u}$ and the view direction $\bm{v}_d \in \R^3$.

There are multiple ways to perform volume rendering with an SDF~\cite{Unisurf, Volsdf, NeuS, HF-NeuS}.
We follow HF-NeuS~\cite{HF-NeuS} and compute the density value $\sigma \in \R_0^+$ by modeling the transparency as the transformed SDF:
\begin{equation}
\sigma (x,y,z) = s\left(\Psi_s\left( {\sdf( x,y,z)} \right) -1 \right)\nabla \sdf( x,y,z) \cdot \bm{v}_d
\label{eq:sigma}
\end{equation}
where $\Psi_s$ is the sigmoid function with scale parameter $s \in \R$ and $\bm{v}_d$ is the viewing direction.
Following NeRF~\cite{Nerf}, the volume rendering integral is approximated via $\alpha$-compositing with $\alpha_i = 1 - \exp\left(-{\sigma_i} \delta_i \right)$, where $\delta_i \in \R_+$ is the distance between adjacent ray samples.
The integrated color values constitute the pixel color for the corresponding ray.

\subsection{Geometric initialization for \tpbackbone}

Prior works (e.g., \cite{Ini_Sdf, NeuS}) showed that proper initialization of the surface can substantially improve its final reconstruction quality.
In our experiments, we confirm this observation and find that \tpbackbone\ converges to a worse solution when naively initialized from random noise (see the Fig.~\apref{fig:add_geoinit} in Appx~\apref{ap:extra-ablations}).
Popular \textit{geometric initialization}~\cite{Ini_Sdf} initializes the parameters of an MLP-based SDF in such a way, that it (approximately) represents a sphere.
Unfortunately, it is designed exclusively for MLPs and thus not directly applicable to grid-based surface representations like tri-planes.
To circumvent this, we develop the following simple technique of adapting geometric initialization for a tri-plane-based SDF.

Our initialization strategy is based on converting a properly initialized MLP into a tri-plane representation.
For this, we take an 8-layer MLP, initialize it with conventional geometric initialization, and then substitute its first 5 layers with tri-planes.
Each $T_{xy}, T_{yz}$, and $T_{xz}$ feature plane is set to values from the MLP by looking up its features in $R \times R$ coordinates of the form $(x, y, 0)$, $(0, y, z)$, and $(x, 0, z)$, respectively.
The remaining 3 layers are used as the MLP head on top of the tri-planes to estimate the distance value (see Sec~\ref{sec:method:tpbackbone}).

\begin{figure*}[!t]
\begin{center}
\includegraphics[width=1.0\textwidth]{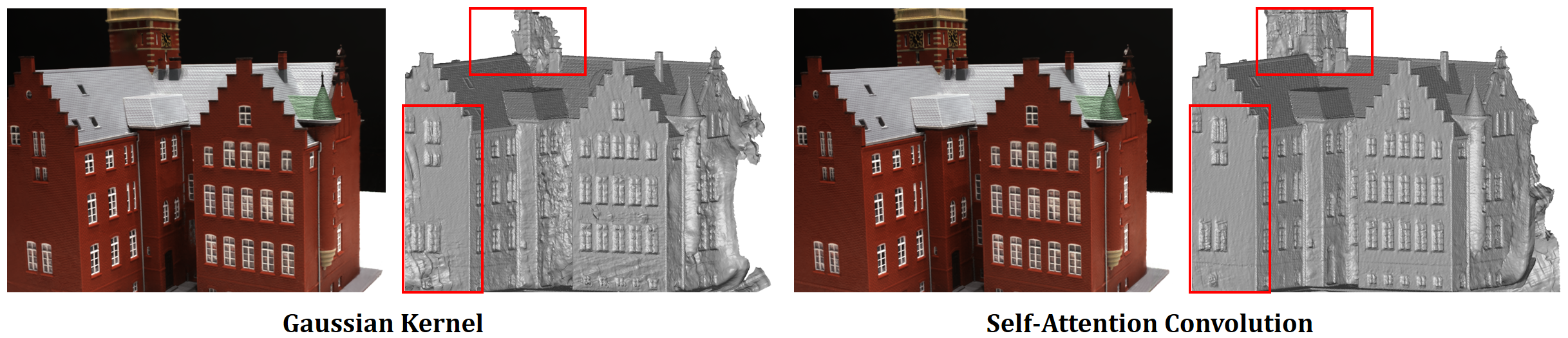}
\end{center}
\caption{Comparing Gaussian kernels with our self-attention kernels. For each method, the left shows the reconstructed image and the right the reconstructed surface.}
\label{fig:kernels}
\end{figure*}

\subsection{Positional Encoding for \tpbackbone}

Traditional positional encoding~\cite{Nerf, NeuS} uses sinusoidal functions to map raw 3D coordinates into multiple different frequencies to make the network capture the characteristics of different frequency scales.
In contrast, tri-planes are an interpolation-based grid representation, which makes it difficult to incorporate similar inductive biases.
To mitigate this, we first derive an implicit function representation as a weighted sum of sinusoidal positional embeddings and then design a method of combining them with tri-planes.

A neural implicit function learns a mapping from the coordinates into the function values.
A continuous unary function with compact support can be expanded as a Fourier series with a frequency scale $M \mapsto \infty$.
In practice, we have only a finite number of frequencies. An MLP can model the coefficients of the Fourier decomposition and also approximate the error caused by finite truncation:
\begin{align}
f\left( x \right) & = {a_0} + \sum\limits_{m=1}^{M} {{a_m}\cos \left( {mx} \right) + } \sum\limits_{m=1}^{M} {{a{_{-m}}}\sin \left( {mx} \right)} \\
& = \mlp(\left\{\cos(mx),\sin(mx)\right\}_{m=1}^M),
\end{align}
where $a_m, a_{-m}$ denote the amplitude coefficients.
We can rewrite the $f(x)$ decomposition as:
\begin{align}
f\left( x \right) & = \sum\limits_{m=-M}^{M} {{a_m} \Theta^x_m}
\end{align}
where $\Theta_t^x$ is an auxiliary variable introduced for brevity:
\begin{equation}\label{eq:theta-def}
\Theta_t^x = \left\{ {\begin{array}{*{20}{c}}
{\cos \left( {tx} \right)}&{t > 0}\\
1&{t = 0}\\
{\sin \left( {tx} \right)}&{t < 0}
\end{array}} \right.
\end{equation}

We can similarly re-write the Fourier series decomposition for an implicit function $f(x,y,z)$ in 3D:
\begin{align}\label{eq:3d-fourier}
f({x,y,z}) &= \sum\limits_{k =  - K}^K {\sum\limits_{n =  - N}^N {\sum\limits_{m =  - M}^M {{a_{mnk}}\Theta _m^x} \Theta _n^y} \Theta _k^z}
\end{align}
where $m$, $n$, and $k$ are the different frequencies for $x, y$, and $z$ axes with the maximum number of frequency scales of $M,N,K \mapsto \infty$, and $a_{mnk}$ denote the corresponding amplitude values.

In the above representation, sine and cosine waves from different dimensions $x,y,z$ are entangled with each other through multiplications.
We want to obtain a representation where each sinusoidal function can be treated individually as an independent input. 
To do this, we perform a series of substitutions.

First, we substitute $\Theta_m^x$ back using its definition~\eqref{eq:theta-def}, we get:
\begin{align}
f & = \sum\limits_{k =  - K}^K {\sum\limits_{n =  - N}^N {\left( { \sum\limits_{m=1}^{M} {{a_{mnk}}\cos \left( {mx} \right)  }  } \right) \Theta _n^y} \Theta _k^z} \\
& + \sum\limits_{k =  - K}^K {\sum\limits_{n =  - N}^N {\left( {  \sum\limits_{m=1}^{M} {{a_{(-m)nk}}\sin \left( {mx} \right)} } \right) \Theta _n^y} \Theta _k^z} \\
& + \sum\limits_{k =  - K}^K {\sum\limits_{n =  - N}^N {\left( {{a_{0nk}} } \right) \Theta _n^y} \Theta _k^z}
\end{align}

The first and second terms can be rewritten as a combination of ${\cos(mx)}$ and ${\sin(mx)}$ by introducing two auxiliary functions $\hat{g}_m(y,z)$ and $\hat{g}'_m(y,z)$:
\begin{align}
\hat{g}_m(y,z) &= \sum\limits_{k =  - K}^K {\sum\limits_{n =  - N}^N { {  {{a_{mnk}} }  }  \Theta _n^y} \Theta _k^z} \\
\hat{g}'_m(y,z) &= \sum\limits_{k =  - K}^K {\sum\limits_{n =  - N}^N { {  {{a_{(-m)nk}} }  }  \Theta _n^y} \Theta _k^z}
\end{align}
Then, we alternatively expand $\Theta _n^y$ and $\Theta _k^z$ in a similar manner to represent the first and second terms of $f(x,y,z)$ as a combination of $\cos(ny), \sin(ny), \cos(kz)$, and $\sin(kz)$.
This yields similar functions $\hat{h}_n$, $\hat{h}'_n$, $\hat{w}_k$ and $\hat{w}'_k$. We provide the detailed derivations for the third terms and yield ${g}_m$, ${g}'_m$, ${h}_n$, ${h}'_n$, ${w}_k$ and ${w}'_k$ for $f(x,y,z)$ in Appx~\apref{ap:derivations}.

As a result, we can approximate the implicit function $f(x,y,z)$ with an MLP with sine/cosine inputs of the following form:
\begin{equation}
f(x,y,z)
\approx \mlp\left( {\left\{ {\begin{array}{*{20}{c}}
{\cos \left( {mx} \right)}\\
{\sin \left( {mx} \right)}\\
{{g_m}\left( {y,z} \right)\cos \left( {mx} \right)}\\
{{g'_m}\left( {y,z} \right)\sin \left( {mx} \right)}\\
{\cos \left( {ny} \right)}\\
{\sin \left( {ny} \right)}\\
{{h_n}\left( {x,z} \right)\cos \left( {ny} \right)}\\
{{h'_n}\left( {x,z} \right)\sin \left( {ny} \right)}\\
{\cos \left( {kz} \right)}\\
{\sin \left( {kz} \right)}\\
{{w_k}\left( {x,y} \right)\cos \left( {kz} \right)}\\
{{w'_k}\left( {x,y} \right)\sin \left( {kz} \right)}
\end{array}} \right\}_{mnk}^{{\bm{flatten}}}} \right)
\label{eq:mlp}
\end{equation}

Such an approximation is directly applicable to our SDF function $\sdf$.
Conventional neural surface reconstruction methods represent the surface entirely through a neural network: $\sdf(x,y,z) = \mlp(\pe(x,y,z))$.
Since the functions $g$, $h$, and $w$ in \eqref{eq:mlp} are all highly nonlinear, the underlying $\mlp$ should be of high capacity to have a low reconstruction error.
Tri-planes can take advantage of the high-capacity features to replace the large MLP backbone network, but this direct replacement poses the following issue. Tri-planes do not carry any inductive biases about positional encoding, and due to the discrete discontinuities of their grid-based representation and the absence of frequency constraints, the tri-plane features will introduce high-frequency noise.
Analyzing Eq.~\eqref{eq:mlp} suggests the following way to integrate positional information into tri-planes.

From \eqref{eq:mlp}, we observe that the coefficients $g$, $h$, and $w$ of positional encoding are consistent with the tri-plane features since these coefficients are all binary functions of the two-dimensional coordinates.
We propose to regard $g$, $h$, and $w$ functions as the tri-plane features and modulate/multiply them with sin and cos functions of different frequencies.
In this way, the output features of tri-planes contain a frequency bound, and the base function should be easier to fit.
This analysis leads to the following parametrization:
\begin{equation}
\sdf(x,y,z) = \mlp\left( {\left[ {\begin{array}{*{20}{c}}
{\pe(x,y,z)}\\
{{T_{xy}}(x,y) \odot \pe(z)}\\
{{T_{yz}}(y,z) \odot \pe(x)}\\
{{T_{xz}}(x,z) \odot \pe(y)}
\end{array}} \right]} \right)
\label{eq:PET}
\end{equation}
where $\odot$ denotes element-wise multiplication.
Such modulation of tri-plane features via conventional positional embeddings endows the grid-based representation with frequency information, which suppresses high-frequency noise.

\label{results}
\begin{table*}[!t]
\caption{
Quantitative results on the NeRF-synthetic dataset. Chamfer distance is on the left, and PSNR is on the right.}
\centering
\tiny
\resizebox{1.0\linewidth}{!}{
\setlength{\tabcolsep}{1.2mm}
{\begin{tabular}{lccccccc|cccccccc} 
\toprule
Method & Chair & Ficus & Lego & Materials & Mic & Ship & \textbf{Mean} & Chair & Ficus & Lego & Materials & Mic & Ship & \textbf{Mean} \\
\midrule
NeRF & 2.12 & 5.17 & 3.05 & 1.51 & 4.77 & 3.54 & 3.36 & \textbf{33.00} & \textbf{30.15} & \textbf{32.54} & 29.62 & 32.91 & \textbf{28.34} & \textbf{31.09} \\
VOLSDF & 1.26 & 1.54 & 2.83 & 1.35 & 3.62 & 2.92 & 2.37 & 25.91 & 24.41 & 26.99 & 28.83 & 29.46 & 25.65 & 26.86 \\
NeuS & 0.74 & 1.21 & 2.35 & 1.30 & 3.89 & 2.33 & 1.97 & 27.95 & 25.79 & 29.85 & 29.36 & 29.89 & 25.46 & 28.05 \\
HF-NeuS & 0.69 & 1.12 & 0.94 & 1.08 & 0.72 & 2.18 & 1.12 & 28.69 & 26.46 & 30.72 & 29.87 & 30.35 & 25.87 & 28.66 \\
\modelname\ (ours) & \textbf{0.65} & \textbf{0.71} & \textbf{0.58} & \textbf{1.05} & \textbf{0.49} & \textbf{1.57} & \textbf{0.84} & 29.57 & 27.39 & 32.40 & \textbf{29.97} & \textbf{33.08} & 26.83 & 29.87 \\
\bottomrule
\end{tabular}
}
}
\label{tab:nerf_table}
\end{table*}   

\begin{figure*}[!t]
\begin{center}
\includegraphics[width=1.0\textwidth]{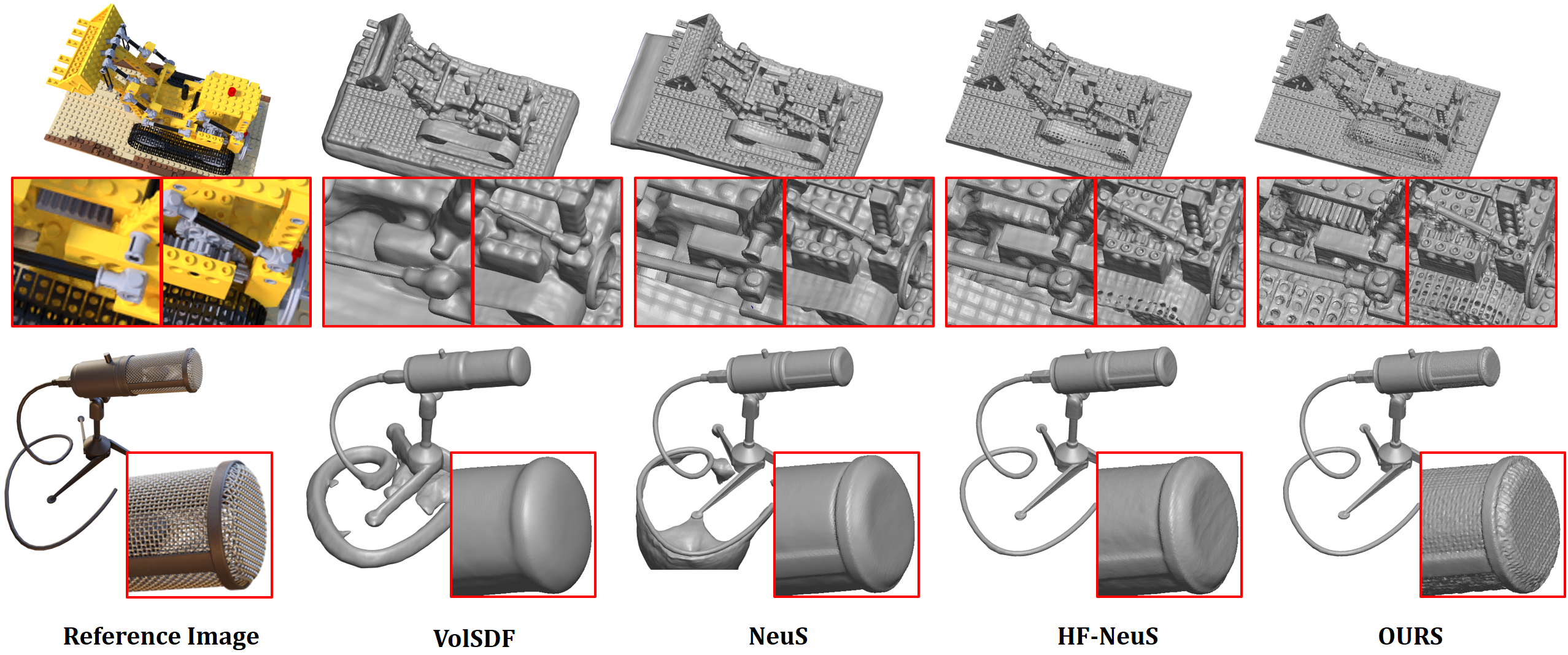}
\end{center}
\caption{Qualitative evaluation on the Lego and Mic models. First column:  reference images. Second to the fifth column: VolSDF, NeuS, HF-NeuS, and OURS.}
\label{fig:nerf_dataset}
\end{figure*}

\subsection{Tri-Planes with Self-Attention}

In practice, one can simply set the feature dimensionality of the tri-planes $n_f$ to the dimension of the positional embedding.
However, in order to better learn the features of different frequencies, we propose to generate tri-plane features with different frequency bands, where each band contains multiple frequency scales.

Since the product in the frequency domain is the convolution in the spatial domain, one can generate multi-frequency tri-plane features by smoothing them with fixed Gaussian kernels.
But when experimenting with this simple technique, we noticed that it smooths features across depth discontinuities, e.g. foreground to background, as depicted in Fig.~\ref{fig:kernels}.
This happens because each feature plane of the tri-plane representation is an orthogonal projection of the 3D space, and foreground features are getting affected by the background features due to the direct convolution on the plane.
This results in the wrong structure of the generated surface, although the rendered image is reasonable.
Therefore, we looked for alternative dynamic convolution operations that could be performed.

Inspired by window self-attention convolution~\cite{WSA} and the Swin Transformer architecture~\cite{SwinTrans}, we propose to use self-attention convolution with different window sizes to generate features in different frequency bands.
We found that using either a sliding window or a shifted window exponentially increases the computational cost of the optimization procedure.
To reduce it, we use a single-layer self-attention convolution and directly divide the tri-plane into regular non-overlapping patches with different window sizes (we use the window sizes of 4, 8, and 16 everywhere unless stated otherwise).
Since the subsequent MLP combines features of different scales, the features across windows also interact with each other.
We take the output features $T^i \in \R^{R \times R \times n_f}$ produced by the $i$-th SAC convolution, and concatenate them all together with the original features $T^0$ to form the final tri-plane representation for four frequency bands as follows:
\begin{equation}
T = \text{concat}\left[T^i\right]_{i=0}^3. \quad T^i=\left\{T^i_{xy},T^i_{yz},T^i_{xz}\right\},
\end{equation}
where $\text{concat}[\cdot]$ is the channel-wise concatenation operation.
We multiply the features of the four frequency bands with the corresponding low-frequency to high-frequency positional encoding.
In this way, we obtain the tri-plane representation with adaptive features for different frequencies, which is well suited for surface reconstruction tasks.

\section{Results}

\subsection{Optimization}
We use two different losses in the training (identical to what has been used in previous work NeuS and HF-NeuS). The first one is the color reconstruction loss. The second is the Eikonal loss~\cite{Eikonal}. We found that total variation regularization~\cite{Tvloss} (TVloss) can also regularize the SDF like Eikonal loss, but Eikonal loss is especially suitable for learning SDF.
Color reconstruction loss is the L1 distance between ground truth colors and the volume rendered colors of sampled pixel set $S$.
\begin{equation}
{\Lcolor} = \frac{1}{|S|}\sum\limits_{s\in S} {{{\left\| {{{\hat C}_s} - {C_s}} \right\|}_1}} 
\end{equation}
Eikonal loss is a regularization loss on sampled point set $I$ that constrains the implicit function and makes the SDF smooth.
\begin{equation}
{\Lreg} = \frac{1}{|I|}\sum\limits_{i\in I} {\left[ {{{\left( {{{\left\| {\nabla \sdf(x_i, y_i, z_i)} \right\|}_2} - 1} \right)}^2}} \right]}
\end{equation}
We employ both loss functions to train our network with a hyperparameter $\lambda$. Note that in all settings we do not provide masks and ignore mask loss in the training.
\begin{equation}
\Loss = {\Lcolor} + \lambda{\Lreg}
\end{equation}

%------------------------------------------------------------------------
\begin{table*}[!t]
\caption{
    Quantitative results on DTU (the header's numbers denote scene IDs). Chamfer distance is on top and PSNR is on the bottom.}
\centering
\tiny
\resizebox{1.0\linewidth}{!}{
    \setlength{\tabcolsep}{0.9mm}
    {\begin{tabular}{lcccccccccccccccc}
	\toprule
	Method & 24 & 37 & 40 & 55 & 63 & 65 & 69 & 83 & 97 & 105 & 106 & 110 & 114 & 118 & 122 & \textbf{Mean} \\
	\midrule
 NeRF   & 1.90 & 1.60 & 1.85 & 0.58 & 2.28 & 1.27 & 1.47 & 1.67 & 2.05 & 1.07 & 0.88 & 2.53 & 1.06 & 1.15 & 0.96 & 1.49 \\
% 	& 
	VOLSDF & 1.14 & 1.26 & 0.81 & 0.49 & 1.25 & 0.70 & 0.72 & 1.29 & 1.18 & 0.70 & 0.66 & 1.08 & 0.42 & 0.61 & 0.55 & 0.86 \\
% 	& 
	NeuS   & 1.00 & 1.37 & 0.93 & 0.43 & 1.10 & 0.65 & \textbf{0.57} & 1.48 & 1.09 & 0.83 & 0.52 & 1.20 & 0.35 & \textbf{0.49} & 0.54 & 0.84 \\
% 	& 
	HF-NeuS   & 0.76 & 1.32 & 0.70 & 0.39 & 1.06 & \textbf{0.63} & 0.63 & \textbf{1.15} & 1.12 & 0.80 & 0.52 & 1.22 & \textbf{0.33} & \textbf{0.49} & 0.50 & 0.77 \\
% 	& 
	\modelname\ (ours)   & \textbf{0.56} & \textbf{0.75} & \textbf{0.68} & \textbf{0.36} & \textbf{0.87} & 0.76 & 0.69 & 1.33 & \textbf{1.08} & \textbf{0.66} & \textbf{0.51} & \textbf{1.04} & 0.34 & 0.51 & \textbf{0.48} & \textbf{0.71} \\
	\midrule
 NeRF   & 26.24 & 25.74 & 26.79 & 27.57 & 31.96 & 31.50 & 29.58 & 32.78 & 28.35 & 32.08 & 33.49 & 31.54 & 31.0 & 35.59 & 35.51 & 30.65 \\
% 	& 
	VOLSDF & 26.28 & 25.61 & 26.55 & 26.76 & 31.57 & 31.50 & 29.38 & 33.23 & 28.03 & 32.13 & 33.16 & 31.49 & 30.33 & 34.90 & 34.75 & 30.38 \\
% 	& 
	NeuS   & 28.20 & 27.10 & 28.13 & 28.80 & 32.05 & 33.75 & 30.96 & 34.47 & 29.57 & 32.98 & 35.07 & 32.74 & 31.69 & 36.97 &  37.07 & 31.97 \\
% 	& 
	HF-NeuS   & 29.15 & 27.33 & 28.37 & 28.88 & 32.89 & \textbf{33.84} & \textbf{31.17} & 34.83 & \textbf{30.06}
 & 33.37 & 35.44 & 33.09 & 32.12 & 37.13 & 37.32 & 32.33 \\
%  & 
 \modelname\ (ours)   & \textbf{30.15} & \textbf{27.69} & \textbf{29.17} & \textbf{29.55} & \textbf{33.78} & 33.65 & 30.96 & \textbf{35.21} & 29.53
 & \textbf{33.43} & \textbf{36.58} & \textbf{33.54} & \textbf{32.34} & \textbf{38.50} & \textbf{37.61} & \textbf{32.78} \\
	\bottomrule
\end{tabular}
}
}
\label{tab:dtu_table}
\end{table*}

\begin{figure*}[!t]
\begin{center}
\includegraphics[width=1.0\textwidth]{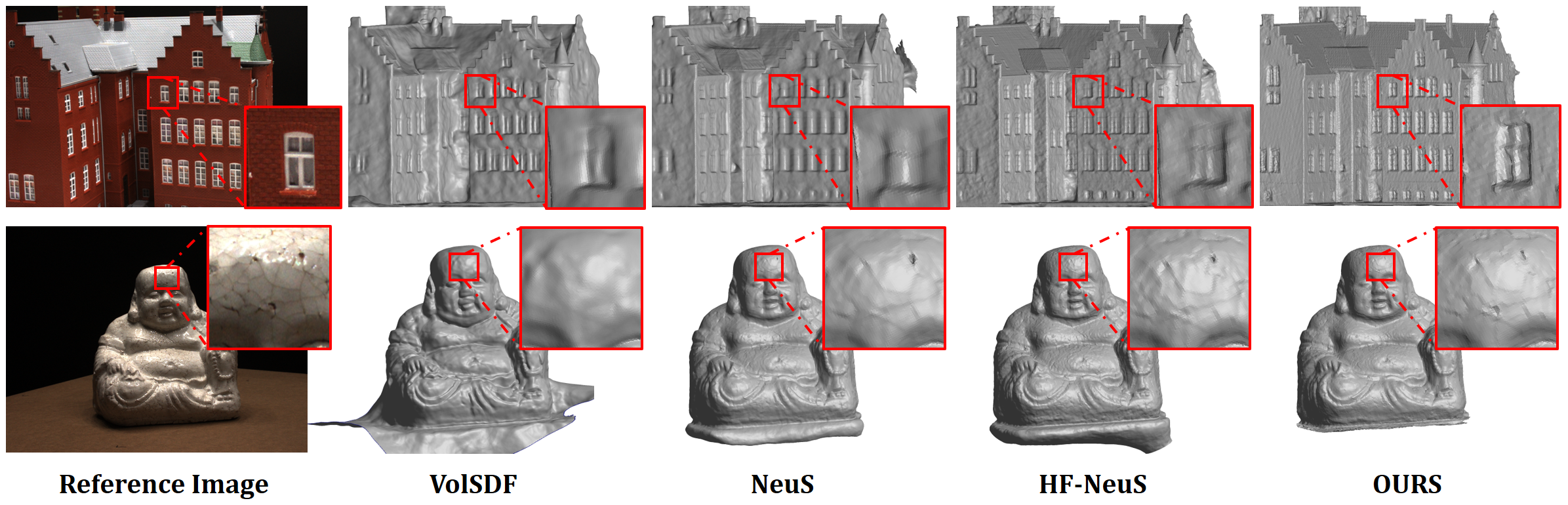}
\end{center}
\caption{Qualitative evaluation on DTU house and Buddha models. First column:  reference images. Second to the fifth column: VolSDF, NeuS, HF-NeuS, and OURS.}
\label{fig:dtu_mvs}
\end{figure*}

\noindent\textbf{Datasets.} 
The NeRF synthetic dataset~\cite{Nerf} contains posed multi-view images of 800 $\times$ 800 resolution with detailed and sharp features. 
The DTU dataset~\cite{DTU} is a real dataset that contains posed multi-view images of 1600 $\times$ 1200 resolution. We select the same 15 models as shown in other works for a fair comparison. The DTU dataset contains non-Lambertian surfaces which are testing for methods sensitive to noise. Besides the DTU dataset, 6 challenging scenes are selected from the NeRF-synthetic dataset.
Ground truth surfaces and camera poses are provided in both datasets.
We also conduct an experiment on other real-world scenes in the Appx~\apref{ap:other-real-data}.

\noindent\textbf{Baselines.}
Four state-of-the-art baselines are considered: 
VolSDF~\cite{Volsdf} embeds an SDF into the density function and employs an error bound by using a sampling strategy. The training time is 12 hours on the DTU dataset.
NeuS~\cite{NeuS} incorporates an SDF into the weighting function and uses sigmoid functions to control the slope of the function. The training time is 16 hours on the DTU dataset.
HF-NeuS~\cite{HF-NeuS} builds on NeuS using offset functions. The training time is 20 hours on the DTU dataset.
NeRF~\cite{Nerf} does not focus on high-fidelity surface reconstruction but high-quality image synthesis, hence producing low-quality surfaces. We include this method for completeness and the training time is 10 hours on the DTU dataset.
Since NeuS and VolSDF compared to older methods and demonstrated better results for surface reconstruction, we do not compare with methods such as IDR~\cite{IDR}, or UNISURF~\cite{Unisurf}. 

\noindent\textbf{Evaluation metrics.}  For the DTU dataset, we follow the official evaluation protocol to evaluate the Chamfer distance. For the NeRF synthetic dataset, we compute the Chamfer distance between the ground truth shape and the reconstructed surface. For completeness, PSNR metric is used to measure the quality of reconstructed images. However, we would like to emphasize that the Chamfer distance is the most important metric for surface reconstruction methods.
% BMVS

\begin{table*}[!t]
\caption{Ablation study results (Chamfer distance) on DTU (the header's numbers denote scene IDs). ``PE'' stands for positional encoding, ``MSA'' --- for multi-scale architecture, ``SAC'' --- self-attention convolutions, ``MPE'' --- modulating positional encoding. We consider tri-planes with PE, MPE and SAC as our full model.}
\centering
\resizebox{1.00\linewidth}{!}{
\begin{tabular}{lcccccccccccccccc}
\toprule
Method & 24 & 37 & 40 & 55 & 63 & 65 & 69 & 83 & 97 & 105 & 106 & 110 & 114 & 118 & 122 & \textbf{Mean} \\

\midrule
% 	& 
TP-NeuS & 1.51 & 1.32 & 1.77 & 0.66 & 1.53 & 1.34 & 1.11 & 1.59 & 1.58 & 0.90 & 0.83 & 1.84 & 0.92 & 0.77 & 0.72 & 1.23 \\
% 	& 
TP-NeuS + PE   & 1.21 & 1.13 & 1.26 & 0.46 & 1.07 & 0.90 & 0.82 & 1.41 & 1.21 & 0.83 & 0.58 & 1.69 & 0.43 & 0.58 & 0.55 & 0.94 \\
% 	& 
~+ MSA & 1.00 & 0.90 & 1.09 & 0.40 & 0.95 & 0.86 & 0.79 & 1.39 & 1.17 & 0.80 & 0.62 & 1.69 & 0.40 & 0.57 & 0.53 & 0.88 \\
% 	& 
~+ SAC & 0.75 & 0.92 & 0.99 & 0.41 & 0.93 & 0.88 & 0.74 & 1.37 & 1.16 & 0.80 & 0.58 & 1.38 & 0.38 & 0.54 & 0.53 & 0.82 \\
% 	& 
~+ MPE & 0.68 & 0.94 & 0.92 & 0.40 & 0.91 & 0.83 & 0.72 & 1.36 & 1.18 & 0.77 & 0.55 & 1.37 & 0.37 & 0.53 & 0.51 & 0.80 \\
% 	& 
~+ MPE + SAC & 0.56 & 0.75 & 0.68 & 0.36 & 0.87 & 0.76 & 0.69 & 1.33 & 1.08 & 0.66 & 0.51 & 1.04 & 0.34 & 0.51 & 0.48 & 0.71  \\
\bottomrule
\end{tabular}

}
\label{tab:ablation_table}
\end{table*}

\noindent\textbf{Implementation details.} We use two MLPs to model the SDF and color function on top of tri-plane features. Each MLP consists of only 3 layers. The hyperparameter for the Eikonal regularization is $\lambda=0.1$. We normalize scenes to fit $L=3.0$. The resolution of each tri-plane is $512 \times 512$. The number of tri-plane feature channels is $n_f=24$. Our three window sizes are set to 4, 8, and 16. We use positional encoding with 8 scales, which means $M = N = K = 8$ and each band contains 2 scales. 
The Adam optimizer with a learning rate $5e{-}4$ is utilized for network training using a single NVIDIA TITAN V100 graphics card. The training time is 9 hours on the DTU dataset, which is faster than all competitors. 
After training, the time for extracting a mesh with 512 grid resolution is 20 seconds and for rendering a 1600x1200 resolution image is around 100 seconds.

\subsection{Comparison}
We first report quantitative comparisons on the NeRF-synthetic dataset~\cite{Nerf}. 
In Table~\ref{tab:nerf_table}, we show Chamfer distance on the left and the PSNR values on the right. The results show that our proposed framework PET-NeuS has the best surface reconstruction quality compared to all other methods. This means that our network has the ability to better preserve local features.
NeRF performs very well in PSNR. The reason could be that NeRF only focuses on reproducing the colors while the SDF-based reconstruction methods loose some of their color-fitting abilities due to the geometric regularization. 
Besides outperforming other baselines in terms of quantitative error,  we also show the visual effect of the improved reconstruction (Fig.~\ref{fig:nerf_dataset}). We find the reconstruction of the bumps and the wheel holes of the Lego model and the grid of the Mic model to be particularly impressive. The reconstructed fine-grained structures are a lot better than what can be achieved with previous work. 

The quantitative results on the DTU dataset~\cite{DTU} are shown in Table~\ref{tab:dtu_table}. We show Chamfer distance on the top and the PSNR values on the bottom. For the Chamfer distance, PET-NeuS surpasses NeuS and VolSDF. Compared with HF-NeuS, PET-NeuS is even better. Our PSNR outperforms all other competitors on the DTU dataset.
The qualitative results compared with other methods are shown in Fig.~\ref{fig:dtu_mvs}. The reconstructed surfaces by PET-NeuS preserve fine-grained details. For instance, the holes between the  eyes of the Buddha and the windows are more obvious.

\subsection{Ablation study}
In Table~\ref{tab:ablation_table}, we conduct an ablation study to analyze the effect of each component.
``\tpbackbone'' refers to just using tri-planes and MLPs to model the SDF: $\sdf(x,y,z) = \mlp([T(x,y,z)])$.
Learning SDFs with positional encoding (``PE'') is denoted as ``\tpbackbone\ + PE'' with $\sdf(x,y,z) = \mlp([\pe(x,y,z), T(x,y,z)])$.
We consider this as our baseline.
``MPE'' means we modulate tri-plane features with positional encoding as in \eqref{eq:PET}. 
``SAC'' refers to generating features with different frequencies using self-attention convolution.
To show the effectiveness of SAC, we compare it against the architecture with multi-scale tri-planes representation, denoted as ``MSA''.
It uses tri-planes for multiple resolutions and is described in the Appx~\apref{ap:method-details}.
We regard ``\tpbackbone\ + PE + MPE + SAC'' as our proposed full architecture PET-NeuS.
We conduct experiments on the DTU dataset quantitatively.
From the results, we can observe that the result of using only ``\tpbackbone\ + PE'' still results in a large geometric error. We believe that this is due to the discretization discontinuities.
Modulating tri-plane features using positional encoding can suppress noise interference.
Using self-attention convolution will match the positional encoding on different frequencies and smooth the noise, which is better than the multi-scale setting. We also provide quantitative results in Fig.~\ref{fig:MA}, in which an improvement in reconstruction quality can be observed.

\section{Conclusion and Limitations}
We propose PET-NeuS, a novel tri-plane based method for multi-view surface reconstruction. By modulating tri-plane features using positional encoding and producing tri-plane features with different frequencies using self-attention convolution, our surface reconstruction can reduce noise interference while maintaining high fidelity. PET-NeuS produces fine-grained surface reconstruction and outperforms other state-of-the-art competitors in qualitative and quantitative comparisons. 
One limitation is that we still require a long computation time. It would be an exciting avenue of future work to improve computation time by one or two orders of magnitude without drastically sacrificing quality.
Another limitation we observed is a trade-off between reconstructing fine details and adding high-frequency noise to otherwise flat surface areas. As we experimented with many versions of our framework, we observed that network architectures that are more expressive to model surface detail tend to be more prone to overfitting and hallucinating details, e.g. in areas of high-frequency changes in light transport. It would be interesting to investigate this trade-off from a theoretical perspective.
Finally, we would like to state that we do not expect a noteworthy negative societal impact due to research on surface reconstruction.

\section{Acknowledgements}
We would like to acknowledge support from the SDAIA-KAUST Center of Excellence in Data Science and Artificial Intelligence, the National Natural Science Foundation of China (62202076), and the Natural Science Foundation of Chongqing (CSTB2022NSCQ-MSX0924).

%%%%%%%%% REFERENCES
{\small
\bibliographystyle{ieee_fullname}
\bibliography{cvpr2023_conference}
}

% ADDING THE SUPP TO REFERENCE ITS CONTENTS FROM THE MAIN TEXT
\appendix
In these supplementary materials, we first provide a derivation for incorporating positional encoding into tri-planes. We then provide ablation studies for geometric initialization, frequency bands, other architectures, and EMD evaluation. We also show the details of self-attention convolution and the multi-scale architecture we compared to. Furthermore, we conduct experiments and show some examples of other real-world scenes. We finally show more qualitative comparisons to supplement the main text.
\newline\newline

\section{Derivations for Positional Encoding Tri-Planes}\label{ap:derivations}
In this section, we derive Eq.~13 in the main text to incorporate positional encodings into tri-planes. Recall the 3D function $f\left( {x,y,z} \right)$ can be expanded as follows.
\begin{align}
f\left( {x,y,z} \right) & = \sum\limits_{k =  - K}^K {{c_k}\left( {x,y} \right)\Theta _k^z} \\ 
& = \sum\limits_{k =  - K}^K {\sum\limits_{n =  - N}^N {{b_{nk}}\left( x \right)\Theta _n^y} \Theta _k^z }  \\
& = \sum\limits_{k =  - K}^K {\sum\limits_{n =  - N}^N {\sum\limits_{m =  - M}^M {{a_{mnk}}\Theta _m^x} \Theta _n^y} \Theta _k^z}
\end{align}
where $m$, $n$, and $k$ are the different frequencies for $x, y, z$ with maximum number of scales $M,N,K \mapsto \infty$ and 
\begin{equation}
\Theta _t^v = \left\{ {\begin{array}{*{20}{c}}
{\cos \left( {tv} \right)}&{t > 0}\\
1&{t = 0}\\
{\sin \left( {tv} \right)}&{t < 0}
\end{array}} \right.
\end{equation}

The idea is to use $\cos(tv)$ and $\sin(tv)$ to represent the function $f\left( {x,y,z} \right)$. We first expand $\Theta _m^x$ to represent $f\left( {x,y,z} \right)$ into the form of ${\cos \left( {mx} \right)}$ and ${\sin \left( {mx} \right)}$ as follows.
\begin{align}
f & = \sum\limits_{k =  - K}^K {\sum\limits_{n =  - N}^N {\left( { \sum\limits_{m=1}^{M} {{a_{mnk}}\cos \left( {mx} \right)  }  } \right) \Theta _n^y} \Theta _k^z} \\
& + \sum\limits_{k =  - K}^K {\sum\limits_{n =  - N}^N {\left( {  \sum\limits_{m=1}^{M} {{a_{(-m)nk}}\sin \left( {mx} \right)} } \right) \Theta _n^y} \Theta _k^z} \\
& + \sum\limits_{k =  - K}^K {\sum\limits_{n =  - N}^N {\left( {{a_{0nk}} } \right) \Theta _n^y} \Theta _k^z}
\end{align}

There are three terms in the equation. The first and second terms can be rewritten as a combination of ${\cos \left( {mx} \right)}$ and ${\sin \left( {mx} \right)}$ as follows.
\begin{align}
& \sum\limits_{k =  - K}^K {\sum\limits_{n =  - N}^N {\left( { \sum\limits_{m=1}^{M} {{a_{mnk}}\cos \left( {mx} \right)  }  } \right) \Theta _n^y} \Theta _k^z} \\
&=  { \sum\limits_{m=1}^{M}  { \left(\sum\limits_{k =  - K}^K {\sum\limits_{n =  - N}^N {{a_{mnk}}\Theta _n^y} \Theta _k^z}\right)\cos \left( {mx} \right)  }  }   \\
& \sum\limits_{k =  - K}^K {\sum\limits_{n =  - N}^N {\left( {  \sum\limits_{m=1}^{M} {{a_{(-m)nk}}\sin \left( {mx} \right)} } \right) \Theta _n^y} \Theta _k^z} \\
&=  {  \sum\limits_{m=1}^{M} {\left(\sum\limits_{k =  - K}^K {\sum\limits_{n =  - N}^N {{a_{(-m)nk}\Theta _n^y} \Theta _k^z}}\right)\sin \left( {mx} \right)} }  
\end{align}

We define $\hat{g}_m(y,z) = \sum\limits_{k =  - K}^K {\sum\limits_{n =  - N}^N { {  {{a_{mnk}} }  }  \Theta _n^y} \Theta _k^z}$ and $\hat{g}'_m(y,z) = \sum\limits_{k =  - K}^K {\sum\limits_{n =  - N}^N { {  {{a_{(-m)nk}} }  }  \Theta _n^y} \Theta _k^z}$. In this way, the function can be expanded in the following form.
\begin{align}
f & = \left( { \sum\limits_{m=1}^{M} {{\hat{g}_m(y,z)}\cos \left( {mx} \right)  }  } \right)  \\
& + \left( {  \sum\limits_{m=1}^{M} {{\hat{g}'_m(y,z)}\sin \left( {mx} \right)} } \right) \\
& + \sum\limits_{k =  - K}^K {\sum\limits_{n =  - N}^N {\left( {{a_{0nk}} } \right) \Theta _n^y} \Theta _k^z}
\end{align}

However, ${\cos \left( {mx} \right)}$ and ${\sin \left( {mx} \right)}$ do not appear in the third term. We therefore would like to find some other way to express the third term using trigonometric functions. We observe that we can alternately expand the other two terms $\Theta _n^y$ and $\Theta _k^z$ in $f\left( {x,y,z} \right)$ of Eq.~3 in the same way and add them together as follows.
\begin{align}
3f & = { \sum\limits_{m=1}^{M} {{\hat{g}_m(y,z)}\cos \left( {mx} \right)  }  }   \\
& +  {  \sum\limits_{m=1}^{M} {{\hat{g}'_m(y,z)}\sin \left( {mx} \right)} }  \\
& + \sum\limits_{k =  - K}^K {\sum\limits_{n =  - N}^N {\left( {{a_{0nk}} } \right) \Theta _n^y} \Theta _k^z} \\
& +  { \sum\limits_{n=1}^{N} {{\hat{h}_n(x,z)}\cos \left( {ny} \right)  }  }   \\
& +  {  \sum\limits_{n=1}^{N} {{\hat{h}'_n(x,z)}\sin \left( {ny} \right)} }  \\
& + \sum\limits_{k =  - K}^K {\sum\limits_{m =  - M}^M {\left( {{a_{m0k}} } \right) \Theta _m^x} \Theta _k^z} \\
& +  { \sum\limits_{k=1}^{K} {{\hat{w}_k(x,y)}\cos \left( {kz} \right)  }  }   \\
& +  {  \sum\limits_{k=1}^{K} {{\hat{w}'_k(x,y)}\sin \left( {kz} \right)} }  \\
& + \sum\limits_{k =  - N}^N {\sum\limits_{n =  - M}^M {\left( {{a_{mn0}} } \right) \Theta _m^x} \Theta _n^y}
\end{align}
where 
\begin{align}
\hat{h}_n(x,z) &= \sum\limits_{k =  - K}^K {\sum\limits_{m =  - M}^M { {  {{a_{mnk}} }  }  \Theta _m^x} \Theta _k^z} \\
\hat{h}'_n(x,z) &= \sum\limits_{k =  - K}^K {\sum\limits_{m =  - M}^M { {  {{a_{m(-n)k}} }  }  \Theta _m^x} \Theta _k^z} \\
\hat{w}_k(x,y) &= \sum\limits_{n =  - N}^N {\sum\limits_{m =  - M}^M { {  {{a_{mnk}} }  }  \Theta _m^x} \Theta _n^y} \\
\hat{w}'_k(x,y) &= \sum\limits_{n =  - N}^N {\sum\limits_{m =  - M}^M { {  {{a_{mn(-k)}} }  }  \Theta _m^x} \Theta _n^y}
\end{align}

We further expend Eq.~17 as follows.
\begin{align}
& \sum\limits_{k =  - K}^K {\sum\limits_{n =  - N}^N {\left( {{a_{0nk}} } \right) \Theta _n^y} \Theta _k^z} \\
&= \sum\limits_{k =  - K}^K {\left( { \sum\limits_{n=1}^{N} {{a_{0nk}}\cos \left( {ny} \right)  }  } \right) \Theta _k^z} \\
&+ \sum\limits_{k =  - K}^K {\left( {  \sum\limits_{n=1}^{N} {{a_{0(-n)k}}\sin \left( {ny} \right)} } \right) \Theta _k^z} \\
&+ \sum\limits_{k =  - K}^K {\left( {{a_{00k}} } \right) \Theta _k^z} \\
&=  { \sum\limits_{n=1}^{N} {\left(\sum\limits_{k =  - K}^K {{a_{0nk}\Theta _k^z}}\right)\cos \left( {ny} \right)  }  }   \\
&+  {  \sum\limits_{n=1}^{N} {\left(\sum\limits_{k =  - K}^K {{a_{0(-n)k}\Theta _k^z}}\right)\sin \left( {ny} \right)}  }  \\
&+ \sum\limits_{k =  - K}^K {\left( {{a_{00k}} } \right) \Theta _k^z} 
\end{align}

Similarly, we can expand Eq.~20, and Eq.~23.
\begin{align}
& \sum\limits_{k =  - K}^K {\sum\limits_{m =  - M}^M {\left( {{a_{m0k}} } \right) \Theta _m^x} \Theta _k^z} \\
&=  { \sum\limits_{k=1}^{K} {\left(\sum\limits_{m =  - M}^M {{a_{m0k}\Theta _m^x}}\right)\cos \left( {kz} \right)  }  }   \\
&+  {  \sum\limits_{k=1}^{K} {\left(\sum\limits_{m =  - M}^M {{a_{m0(-k)}\Theta _m^x}}\right)\sin \left( {kz} \right)}  }  \\
&+ \sum\limits_{m =  - M}^M {\left( {{a_{m00}} } \right) \Theta _m^x} 
\end{align}
and 
\begin{align}
& \sum\limits_{k =  - N}^N {\sum\limits_{n =  - M}^M {\left( {{a_{mn0}} } \right) \Theta _m^x} \Theta _n^y} \\
&=  { \sum\limits_{m=1}^{M} {\left(\sum\limits_{n =  - N}^N {{a_{mn0}\Theta _n^y}}\right)\cos \left( {mx} \right)  }  }   \\
&+  {  \sum\limits_{m=1}^{M} {\left(\sum\limits_{n =  - N}^N {{a_{(-m)n0}\Theta _n^y}}\right)\sin \left( {mx} \right)}  }  \\
&+ \sum\limits_{n =  - N}^N {\left( {{a_{0n0}} } \right) \Theta _n^y} 
\end{align}

Therefore the function can be rewritten as follows.
\begin{align}
3f & = { \sum\limits_{m=1}^{M} {{\left(\hat{g}_m(y,z) + \sum\limits_{n =  - N}^N {{a_{mn0}\Theta _n^y}}\right)}\cos \left( {mx} \right)  }  }   \\
& +  {  \sum\limits_{m=1}^{M} {{\left(\hat{g}'_m(y,z) + \sum\limits_{n =  - N}^N {{a_{(-m)n0}\Theta _n^y}}\right)}\sin \left( {mx} \right)} }  \\
& +  { \sum\limits_{n=1}^{N} {{\left(\hat{h}_n(x,z) + \sum\limits_{k =  - K}^K {{a_{0nk}\Theta _k^z}}\right)}\cos \left( {ny} \right)  }  }   \\
& +  {  \sum\limits_{n=1}^{N} {{\left(\hat{h}'_n(x,z) + \sum\limits_{k =  - K}^K {{a_{0(-n)k}\Theta _k^z}}\right)}\sin \left( {ny} \right)} }  \\
& +  { \sum\limits_{k=1}^{K} {{\left(\hat{w}_k(x,y) + \sum\limits_{m =  - M}^M {{a_{m0k}\Theta _m^x}}\right)}\cos \left( {kz} \right)  }  }   \\
& +  {  \sum\limits_{k=1}^{K} {{\left(\hat{w}'_k(x,y) + \sum\limits_{m =  - M}^M {{a_{m0(-k)}\Theta _m^x}}\right)}\sin \left( {kz} \right)} }  \\
&+ \sum\limits_{m =  - M}^M {\left( {{a_{m00}} } \right) \Theta _m^x} \\
&+ \sum\limits_{n =  - N}^N {\left( {{a_{0n0}} } \right) \Theta _n^y} \\
&+ \sum\limits_{k =  - K}^K {\left( {{a_{00k}} } \right) \Theta _k^z} 
\end{align}

Now we ignore the constant coefficient 3 and the function $f$ can be rewritten by the combination of positional encoding, which can be learned by an MLP network as follows.
\begin{equation}
f\left( {x,y,z} \right) 
= MLP\left( {\left\{ {\begin{array}{*{20}{c}}
{\cos \left( {mx} \right)}\\
{\sin \left( {mx} \right)}\\
{{g_m}\left( {y,z} \right)\cos \left( {mx} \right)}\\
{{{g'}_m}\left( {y,z} \right)\sin \left( {mx} \right)}\\
{\cos \left( {ny} \right)}\\
{\sin \left( {ny} \right)}\\
{{h_n}\left( {x,z} \right)\cos \left( {ny} \right)}\\
{{{h'}_n}\left( {x,z} \right)\sin \left( {ny} \right)}\\
{\cos \left( {kz} \right)}\\
{\sin \left( {kz} \right)}\\
{{w_k}\left( {x,y} \right)\cos \left( {kz} \right)}\\
{{{w'}_k}\left( {x,y} \right)\sin \left( {kz} \right)}
\end{array}} \right\}_{mnk}^{{\bf{flatten}}}} \right)
\label{ap:eq:mlp}
\end{equation}
where 
\begin{align}
& {g_m}\left( {y,z} \right) = \hat{g}_m(y,z) + \sum\limits_{n =  - N}^N {{a_{mn0}\Theta _n^y}} \\
& {{g'}_m}\left( {y,z} \right) = \hat{g}'_m(y,z) + \sum\limits_{n =  - N}^N {{a_{(-m)n0}\Theta _n^y}} \\
& {h_n}\left( {x,z} \right) = \hat{h}_n(x,z) + \sum\limits_{k =  - K}^K {{a_{0nk}\Theta _k^z}} \\
& {{h'}_n}\left( {x,z} \right) = \hat{h}'_n(x,z) + \sum\limits_{k =  - K}^K {{a_{0(-n)k}\Theta _k^z}} \\
& {w_k}\left( {x,y} \right) = \hat{w}_k(x,y) + \sum\limits_{m =  - M}^M {{a_{m0k}\Theta _m^x}} \\
& {{w'}_k}\left( {x,y} \right) = \hat{w}'_k(x,y) + \sum\limits_{m =  - M}^M {{a_{m0(-k)}\Theta _m^x}}
\end{align}

\begin{figure*}[t]
\begin{center}
\includegraphics[width=1.0\textwidth]{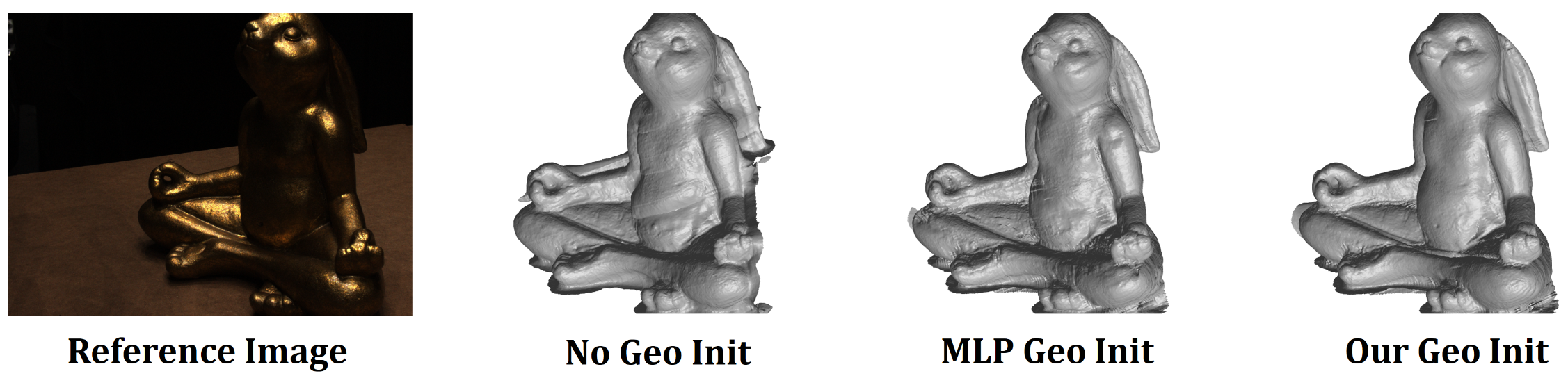}
\end{center}
\caption{Qualitative evaluation for geometric initialization. First column:  reference images. Second to the fourth column: no geometric initialization, geometric initialization for MLP, and our geometric initialization.}
\label{fig:add_geoinit}
\end{figure*}

\begin{table}[b]
\vspace{-0.5cm}
\caption{Ablations for geometric initialization. }
\vspace{-0.3cm}
\centering 
\begin{tabular}{lcccc}
\toprule
Method & 1 & 2 & 3 & \textbf{Mean} \\

\midrule
% 	& 
No Geo Init  & 1.86 & 1.79 & 1.92 & 1.86 \\
% 	& 
MLP Geo Init  & 1.39 & 1.29 & 1.33 & 1.34 \\
% 	& 
Our Geo Init & 1.04 & 1.03 & 1.09 & 1.05 \\

\bottomrule
\end{tabular}
\label{tab:geoinit}
\end{table}

\section{Additional Ablations}\label{ap:extra-ablations}
\subsection{Geometric Initialization}
In this section, we provide an ablation study for geometric initialization. We compare our geometric initialization method (Our Geo Init) with two different settings. The first one is to initialize tri-planes from random noise and the 3-layer MLP with standard geometric initialization~\cite{Ini_Sdf} (MLP Geo Init). The random noise is from a normal distribution 
$\mathcal{N}\left(0, 1\right)$. In the second setting, the standard geometric initialization is not used during the initialization stage (No Geo Init). We show a representative example in Fig.~\ref{fig:add_geoinit} to compare the difference. We can observe that the method without geometry initialization causes the foreground and background to stick together and inconsistent reconstruction results on the belly. Using random noise to initialize tri-planes and standard geometric initialization to initialize MLP results in high-frequency details that are artifacts on the generated surface model. We also show a comparison of chamfer distance in \tabref{tab:geoinit}. We run three times for each setting and take the average as a metric to evaluate the different initialization methods. Our geometric initialization leads to more consistent surface reconstruction results.

\begin{table*}[t]
\caption{Additional ablations and EMD evaluation on DTU. }
\vspace{-0.3cm}
\centering
\tiny
\resizebox{1.00\linewidth}{!}{
	\setlength{\tabcolsep}{0.8mm}
	{\begin{tabular}{lcccccccccccccccc}
\toprule
Method & 24 & 37 & 40 & 55 & 63 & 65 & 69 & 83 & 97 & 105 & 106 & 110 & 114 & 118 & 122 & \textbf{Mean} \\

\midrule
% 	& 
~+ MSA + SAC (2+2 FB)  & 0.72 & 0.89 & 1.05 & 0.38 & 0.92 & 0.86 & 0.81 & 1.40 & 1.17 & 0.78 & 0.60 & 1.55 & 0.41 & 0.55 & 0.50 & 0.84 \\
% 	& 
~+ MSA + MPE (4FB)  & 0.98 & 0.86 & 0.82 & 0.38 & 0.90 & 0.82 & 0.74 & 1.36 & 1.16 & 0.75 & 0.55 & 1.61 & 0.35 & 0.53 & 0.50 & 0.82 \\
% 	& 
\midrule
~+ SAC + MPE (2FB) & 0.59 & 0.78 & 0.70 & 0.36 & 0.89 & 0.80 & 0.75 & 1.35 & 1.12 & 0.68 & 0.53 & 1.10 & 0.34 & 0.51 & 0.50 & 0.73 \\
% 	& 
~+ SAC + MPE (4FB) & 0.56 & 0.75 & 0.68 & 0.36 & 0.87 & 0.76 & 0.69 & 1.33 & 1.08 & 0.66 & 0.51 & 1.04 & 0.34 & 0.51 & 0.48 & \textbf{0.71}  \\
% 	& 
~+ SAC + MPE (6FB) & 0.64 & 0.76 & 0.71 & 0.36 & 0.87 & 0.83 & 0.76 & 1.36 & 1.13 & 0.67 & 0.54 & 1.07 & 0.34 & 0.52 & 0.51 & 0.74  \\
% 	& 
\midrule
\multicolumn{17}{c}{Earth Mover Distance (EMD) evaluation on DTU} \\
\toprule
~ NeuS & 1.03 & 1.08 & 0.85 & 0.96 & 1.22 & 0.80 & 0.84 & 0.98 & 1.01 & 0.88 & 0.71 & 0.87 & 0.75 & 0.80 & 0.82 & 0.91  \\
% 	& 
~ Ours & 0.87 & 0.92 & 0.71 & 0.93 & 1.01 & 0.85 & 0.85 & 0.95 & 1.02 & 0.84 & 0.70 & 0.82 & 0.74 & 0.80 & 0.81 & 0.85  \\
\bottomrule
\end{tabular}
}
}
\label{tab:new_exper}
\end{table*}

\subsection{Frequency Bands}
We conduct an ablation study to investigate the impact of different frequency bands. 
\tabref{tab:new_exper} provides Chamfer distance results for 2, 4, and 6 frequency bands with the window sizes of \{8\}, \{4,8,16\}, and \{1,2,4,8,16\}, respectively. Four frequency bands performed the best in our experiments.

\subsection{Other Combination of Architectures}
We run the experiment for MSA+SAC and MSA+MPE and report the results in \tabref{tab:new_exper}. To construct four frequency bands (4FB) for MSA+SAC, we use two-resolution triplanes for MSA and apply two-frequency-band SAC on each tri-plane. 
Compared with SAC+MPE (4FB) in \tabref{tab:new_exper}, SAC+MPE method is superior to MSA+SAC and MSA+MPE.

\subsection{EMD evaluation}
We follow the very recent SotA methods (e.g., NeuS, VolSDF, IDR), which use \emph{only} Chamfer distance to evaluate the surface quality.
For additional EMD evaluation, we compare our method with NeuS and provide the results in \tabref{tab:new_exper}.
For each method, we evenly sampled 10K points on the reconstructed surface to obtain the point cloud.
The results are consistent with Chamfer distance, and our method outperforms NeuS on DTU by 7\% on average.

\begin{figure*}[t]
   \centering
   \includegraphics[width=1.0\textwidth]{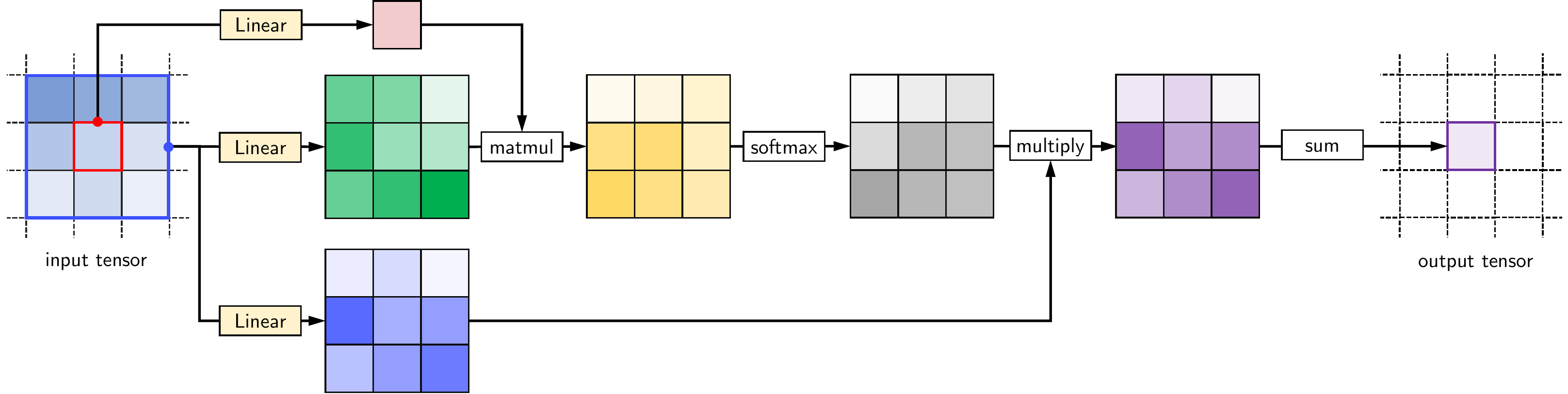}
   \vskip -0.1cm
   \caption{Self-Attention Convolution for a window size $k=3$.}
   \vskip -0.1cm
   \label{fig:SAC}
\end{figure*}

\section{Method Details}\label{ap:method-details}
\subsection{Multi-Scale Architecture}
Here, we discuss the details of the multi-scale architecture we compared to in the main text. One general strategy to achieve multi-scale behavior is to use a multi-resolution data structure.
Therefore, it was our idea to use tri-planes with different resolutions. 
We use Tri-planes with four different resolutions in the multi-scale architecture to mimic four frequency bands in the SAC (self-attention convolution) structure for a fair comparison. For the highest resolution tri-plane representing high frequency, we used the same resolution as SAC, i.e. 512 dimensions. However, for the low-frequency tri-planes, we use three different tri-planes with different resolutions, namely 256$\times$256, 128$\times$128, and 64$\times$64. We then concatenate features generated by these tri-planes to obtain multi-scale tri-plane features. We show the results in Table 3 in the main text and find that our results using the self-attention convolution outperform the results using the multi-scale architecture. An intuitive explanation is that the smoothing kernel can deal with the influence of noise more effectively.
\subsection{Self-Attention Convolution}
We use the vanilla SAC mechanism from \cite{WSA} and its detailed diagram is provided in \figref{fig:SAC}.
Larger window sizes represent lower frequency and smaller window sizes represent higher frequency.
These window sizes are used as different kernels in the self-attention convolution and the corresponding convolved features are produced. 
Adding the original tri-plane features, we can get four triplane features with frequencies as our output feature maps.

\section{Other Real-world Examples}\label{ap:other-real-data}
\begin{figure*}[!t]
\begin{center}
\includegraphics[width=1.0\textwidth]{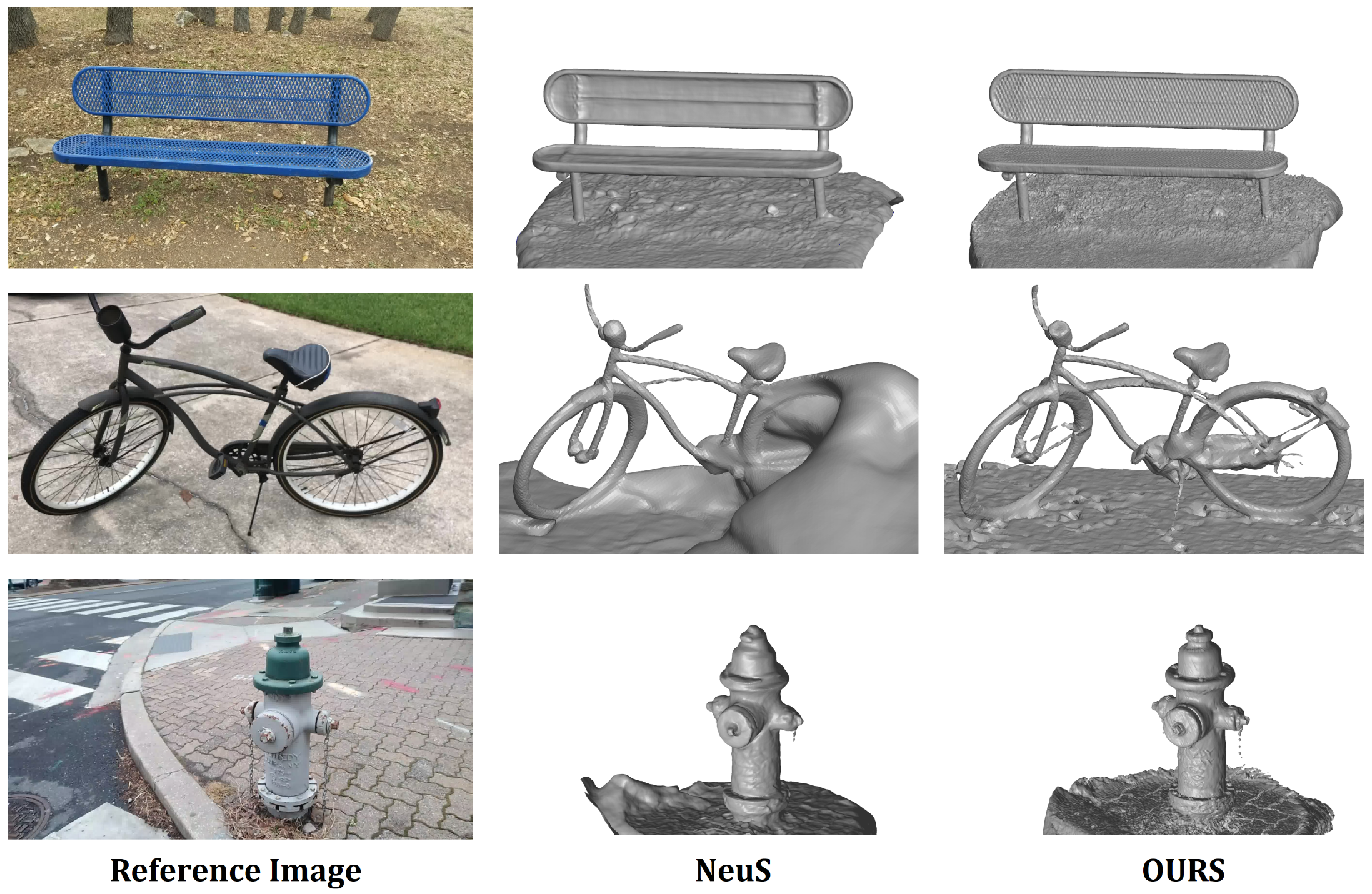}
\end{center}
\caption{Qualitative evaluation on CO3D dataset. First column:  reference images. Second to the third column: NeuS and \modelname.}
\label{fig:add_co3d}
\end{figure*}

We conduct an experiment on another real-world dataset namely CO3D~\cite{co3d}. This dataset provides data for real-world scenes. Many scenes are captured with portable devices and camera poses are extracted using COLMAP~\cite{Structure}. 3D reconstruction methods from datasets with this acquisition method are more general, but also introduce greater noise and challenges. We select some representative examples (bench, bicycle, and hydrant) and show the comparison with NeuS~\cite{NeuS} on these examples in Fig.~\ref{fig:add_co3d}. Experimental results show that our method can reconstruct detailed features, such as the net structure of a bench, the rear wheel of a bicycle, and the rivets of a fire hydrant.

\section{Additional Qualitative Comparisons}
\begin{figure*}[ht]
\begin{center}
\includegraphics[width=1.0\textwidth]{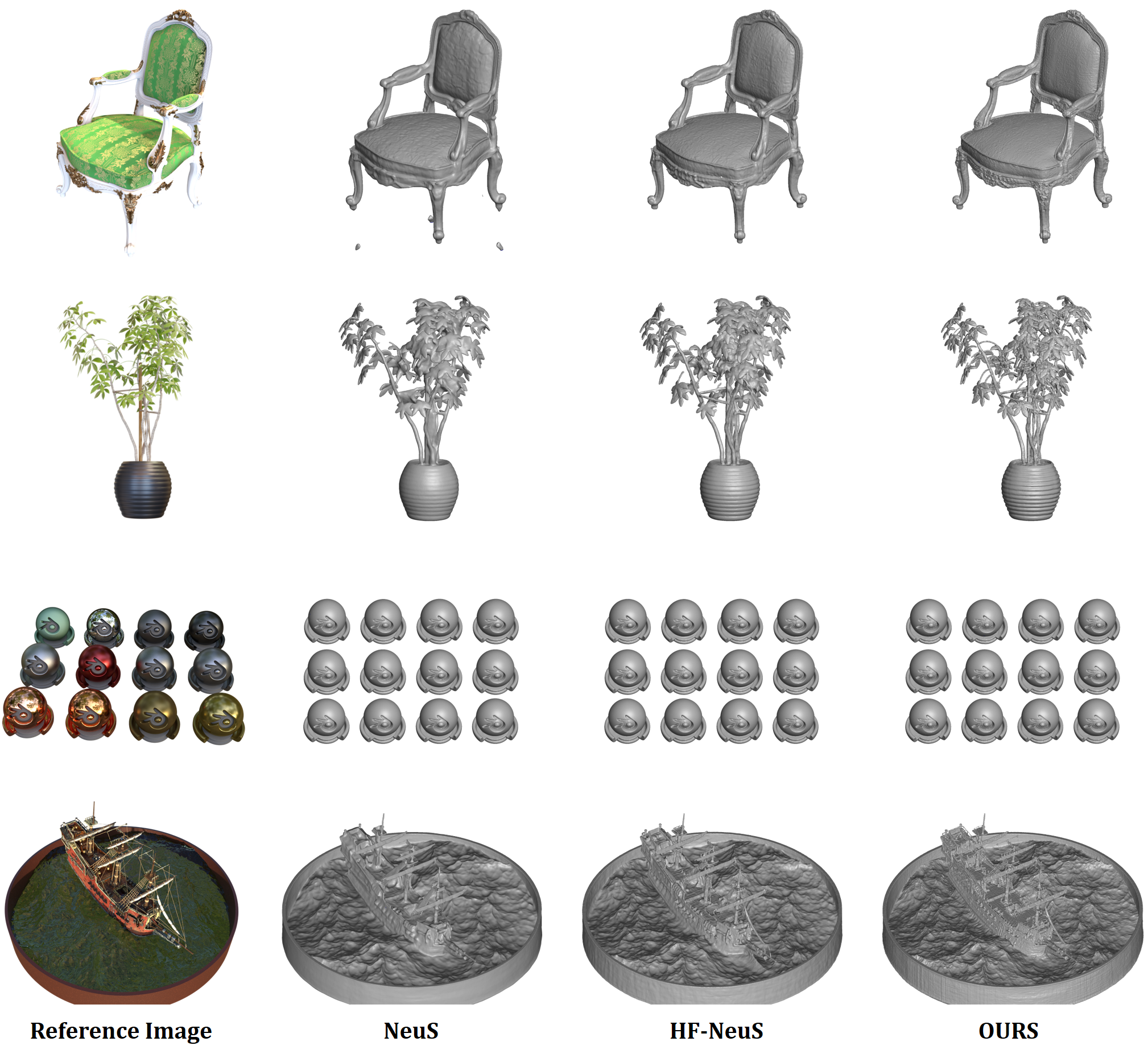}
\end{center}
\caption{Qualitative evaluation on the NeRF synthetic dataset. First column:  reference images. Second to the fourth column: NeuS, HF-NeuS, and \modelname.}
\label{fig:add1}
\end{figure*}

\begin{figure*}[!t]
\begin{center}
\includegraphics[width=0.77\textwidth]{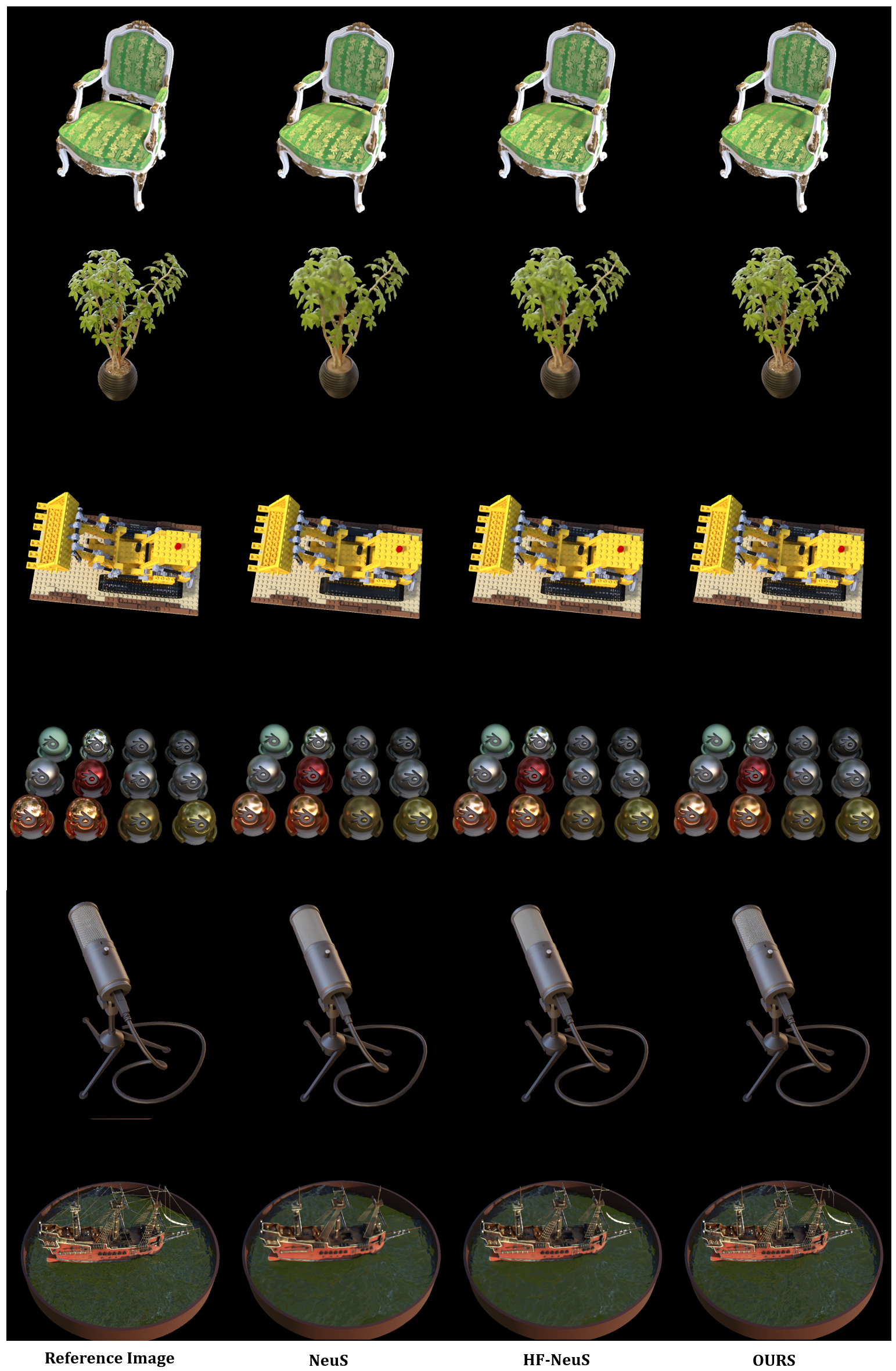}
\end{center}
\caption{Qualitative evaluation on NeRF synthetic dataset. First column:  reference images. Second to the fourth column: the generated images from NeuS, HF-NeuS, and \modelname.}
\label{fig:add3}
\end{figure*}

\begin{figure*}[!t]
\begin{center}
\includegraphics[width=0.92\textwidth]{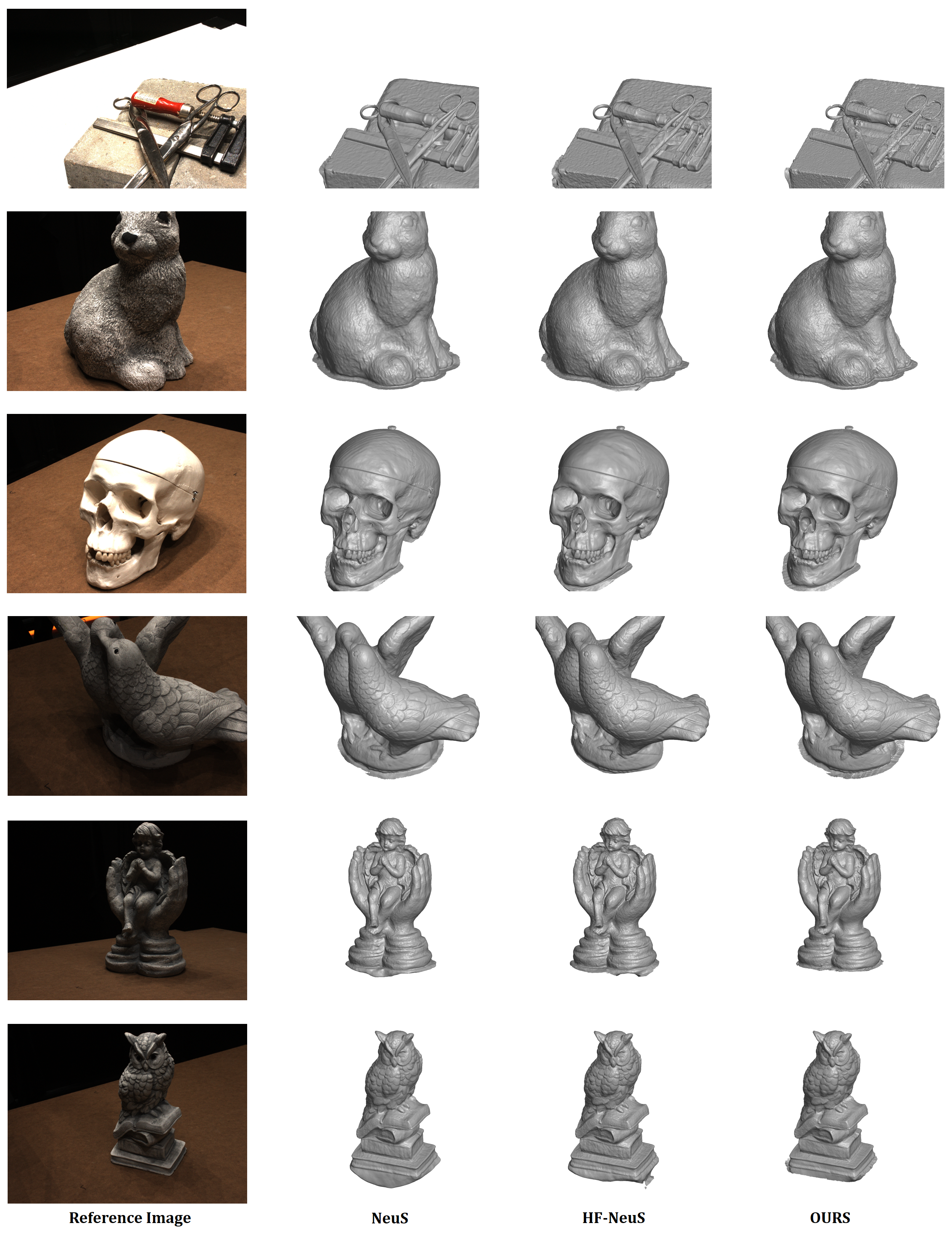}
\end{center}
\caption{Qualitative evaluation on DTU dataset. First column:  reference images. Second to the fourth column: NeuS, HF-NeuS, and \modelname.}
\label{fig:add2}
\end{figure*}

\begin{figure*}[!t]
\begin{center}
\includegraphics[width=1.0\textwidth]{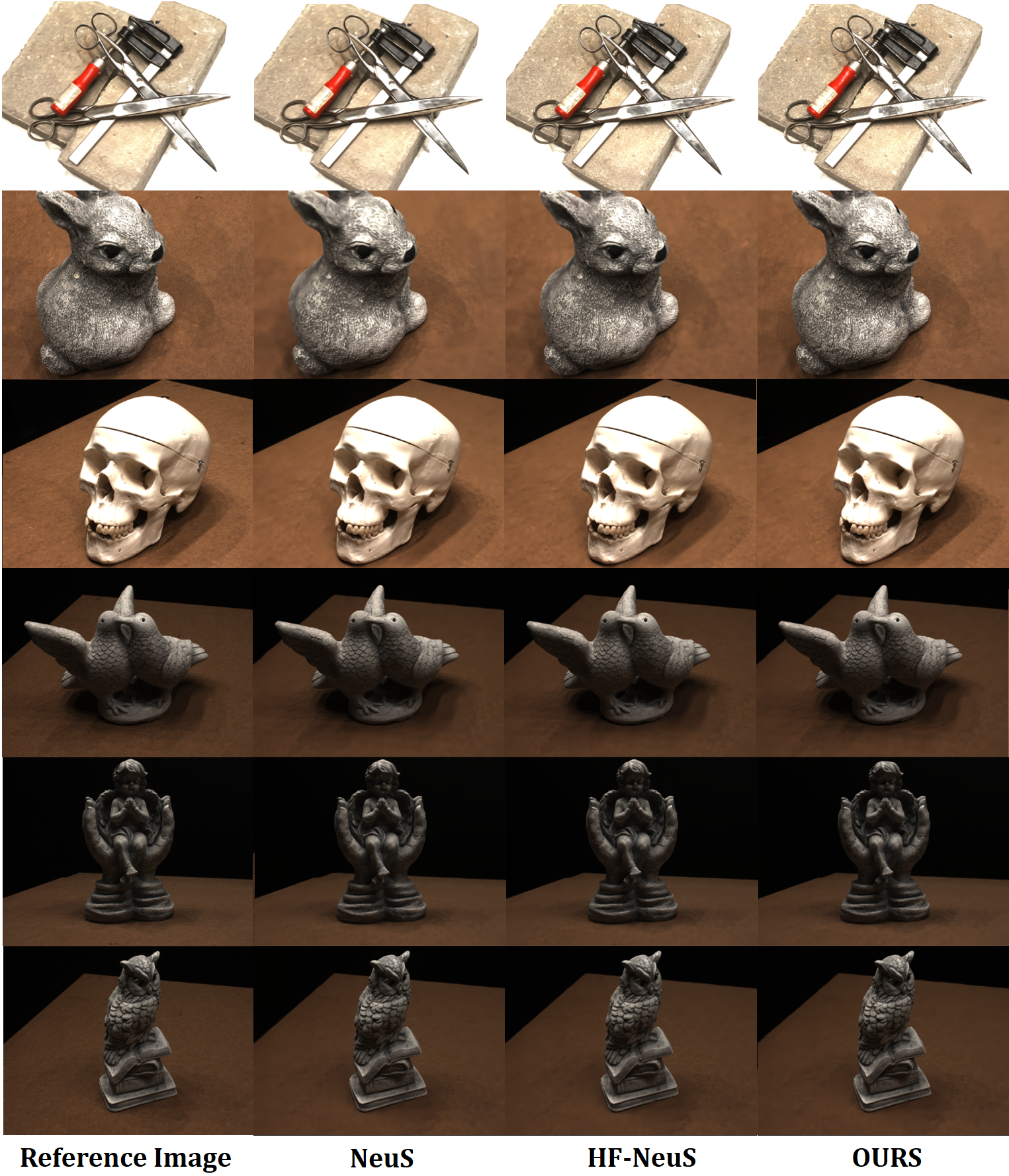}
\end{center}
\caption{Qualitative evaluation on DTU dataset. First column:  reference images. Second to the fourth column: the generated images from NeuS, HF-NeuS, and \modelname.}
\label{fig:add4}
\end{figure*}

In this section, we provide more qualitative comparisons with NeuS~\cite{NeuS} and HF-NeuS~\cite{HF-NeuS} for surfaces and images on the NeRF synthetic dataset (Fig.~\ref{fig:add1} and Fig.~\ref{fig:add3}) and the DTU dataset (Fig.~\ref{fig:add2} and Fig.~\ref{fig:add4}).

\end{document}